\documentclass[sigconf,natbib=false]{acmart}
\usepackage{ascmac}
\usepackage{multirow}
\usepackage{url}
\usepackage[bottom]{footmisc} 

\AtBeginDocument{%
  }

\setcopyright{acmlicensed}
\copyrightyear{2018}
\acmYear{2018}
\acmDOI{XXXXXXX.XXXXXXX}

\acmConference[Conference acronym 'XX]{Make sure to enter the correct
  conference title from your rights confirmation emai}{June 03--05,
  2018}{Woodstock, NY}
\acmISBN{978-1-4503-XXXX-X/18/06}

\RequirePackage[
  datamodel=acmdatamodel,
  style=acmnumeric,
  ]{biblatex}

\begin{document}

\title{News Recommendation with \\ Category Description by a Large Language Model}


\author{Yuki Yada}
\affiliation{%
    \institution{Waseda University}
    \city{Tokyo}
    \country{Japan}}
\email{yada_yuki@yama.info.waseda.ac.jp}

\author{Hayato Yamana}
\affiliation{%
    \institution{Waseda University}
    \city{Tokyo}
    \country{Japan}}
\email{yamana@yama.info.waseda.ac.jp}


\begin{abstract}
    Personalized news recommendations are essential for online news platforms to assist users in discovering news articles that match their interests from a vast amount of online content. Appropriately encoded content features, such as text, categories, and images, are essential for recommendations. Among these features, news categories, such as \textit{tv-golden-globe}, \textit{finance-real-estate}, and \textit{news-politics}, play an important role in understanding news content, inspiring us to enhance the categories' descriptions. In this paper, we propose a novel method that automatically generates informative category descriptions using a large language model (LLM) without manual effort or domain-specific knowledge and incorporates them into recommendation models as additional information. In our comprehensive experimental evaluations using the MIND dataset, our method successfully achieved 5.8\% improvement at most in AUC compared with baseline approaches without the LLM's generated category descriptions for the state-of-the-art content-based recommendation models including NAML, NRMS, and NPA. These results validate the effectiveness of our approach. The code is available at \url{https://github.com/yamanalab/gpt-augmented-news-recommendation}.
\end{abstract}


\received{20 February 2007}
\received[revised]{12 March 2009}
\received[accepted]{5 June 2009}

\maketitle

\section{Introduction}

In recent years, online news platforms have gained widespread popularity, making it commonplace for users to consume news articles on their mobile devices. These platforms deliver a large amount of news content to users daily, leading to an information overload problem, making it challenging to discover articles that align with user's interests. Therefore, personalized news recommendations have become essential for online news platforms.

\begin{figure}[b]
    \includegraphics[width=\linewidth]{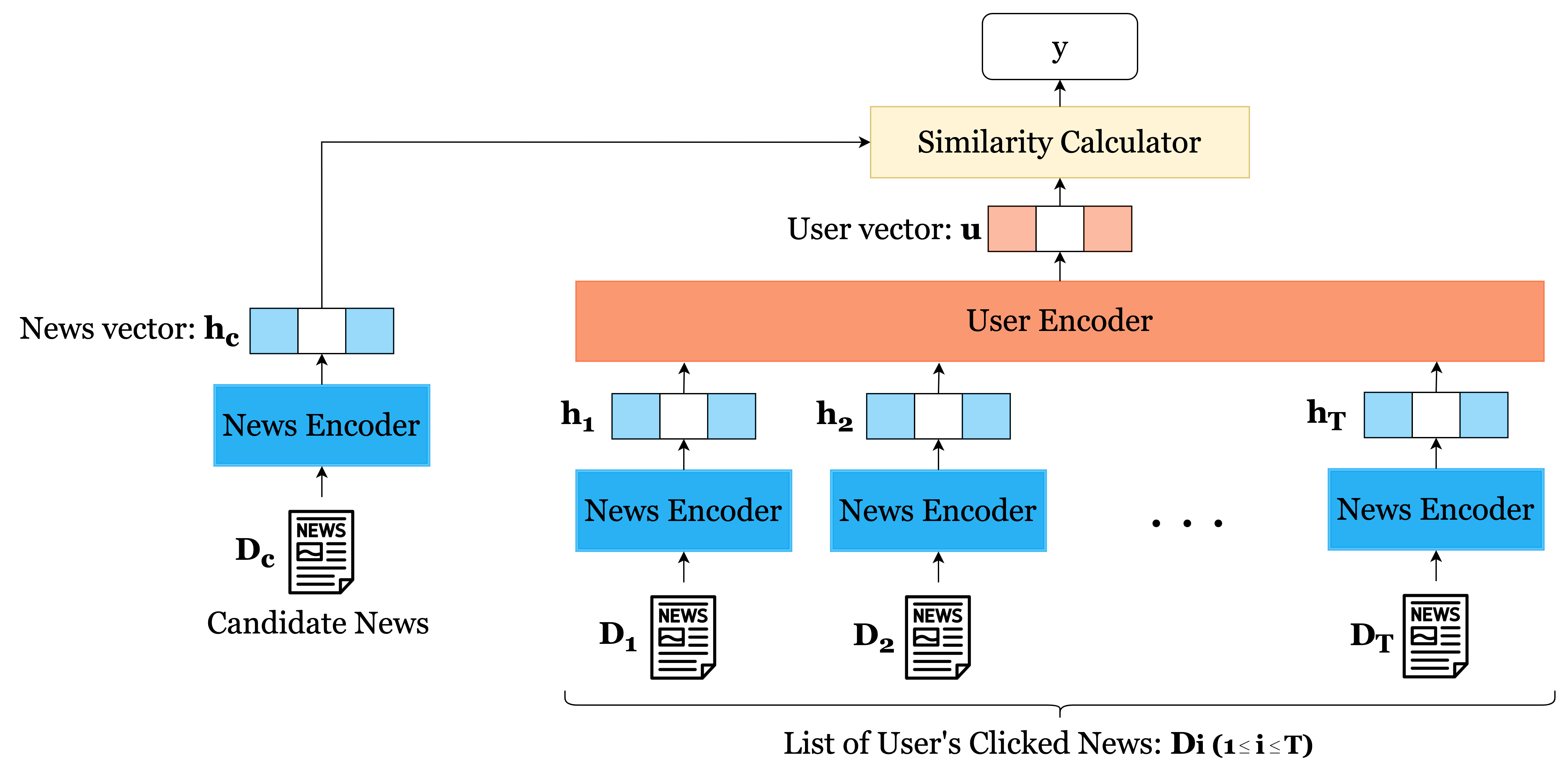}
    \caption{Neural News Recommendation Model}
    \label{fig:newsrec_overview}
\end{figure}

Many proposed news recommendation models use deep neural networks, which achieve high performance \cite{lstur,nrecsurvey2023,naml,npa,nrms,nrecsurvey2021}. As shown in Figure \ref{fig:newsrec_overview}, most neural news recommendation models adopt a common approach that consists of three core modules:

\begin{enumerate}
    \item News encoder: it generates a news vector that captures the semantic information from the news content.
    \item User encoder: it generates a user vector that captures the user preferences based on browsing history.
    \item Similarity calculator: it computes the similarity score between the news vector and the user vector, estimating that news items with higher similarity scores are more likely to match the user’s interests.
\end{enumerate}

Previous studies have focused on investigating which neural network structures to apply in the user and news encoder modules, resulting in various proposed architectures for news recommendation models. Recent methods incorporate pre-trained language models (PLMs) into the news encoder to learn news representations. These PLM-based approaches have achieved high performance \cite{once,nrecsurvey2023,reduceCross,nrecsurvey2021,plmnr}.

Simultaneously, a crucial source for understanding news content is the news category. For example, the MIND \cite{mind} dataset, a well-known dataset in the news recommendation field, contains news categories such as \textit{tv-golden-globes}. The naïve way to use categories in recommendation models is to adopt predefined templates (e.g., \textit{The news category is \{category\}}) followed by combining them with original news recommendation models. However, these templates are too generic to be applied to all news categories, resulting in insufficient information. Constructing detailed descriptions for each news category and using them as input can be beneficial for enabling recommendation models to recognize news content accurately.

However, manually constructing detailed descriptions for each
news category is costly. To address this issue, we propose automatically generating descriptive text for news categories using large language models (LLMs). Since LLMs are pre-trained on vast amounts of text data and have extensive knowledge across diverse topics, the LLM-generated detailed descriptions are expected to enhance the recommendation performance. The contributions of this paper are as follows:

\begin{enumerate}
    \item We first propose the adoptation of LLM-generated news category descriptions to enhance news recommendations.
    \item Comprehensive experimental evaluations on the MIND dataset confirmed that our proposed method consistently outperforms baseline approaches across multiple recommendation models, achieving up to 5.8\% improvement in AUC compared to methods without category descriptions.
\end{enumerate}

The rest of this paper is organized as follows. In Section \ref{sec:related}, we reviews state-of-the-art news recommendations and use of  LLM-generated text. We propose a new recommendation method enhanced by an LLM in Section \ref{sec:method}, followed by evaluations in Section \ref{sec:exp}. In Section \ref{sec:limit}, we discuss the limitations of the proposed method. Finally, In Section \ref{sec:conclusion}, we conclude this paper.

\section{Related Work}\label{sec:related}

In this section, we review related work on news recommendations and the use of LLM-generated text.

\subsection{News Recommendations}

News recommendations have been extensively studied through various models \cite{nrecsurvey2023,nrecsurvey2021}. Neural network-based models have achieved state-of-the-art performance by encoding news content and user preferences to capture the complex interactions between users and news items \cite{lstur,mccm,naml,npa,nrms}. For instance, NRMS \cite{nrms} employs multi-head attention to acquire news and user vectors, while LSTUR \cite{lstur} adopts GRU for to obtain user vectors. Other models, such as MCCM \cite{mccm}, NAML \cite{naml}, and NPA \cite{npa}, leverage CNN and attention mechanisms to generate user and news vectors.

Particularly, the models adopting PLMs, such as BERT \cite{bert} and RoBERTa \cite{roberta}, to obtain news vectors have achieved notably high performance \cite{once,nrecsurvey2023,reduceCross,nrecsurvey2021,plmnr,mmrec}. For instance, Wu et al. \cite{plmnr} proposed a framework incorporating PLMs into news recommendations, demonstrating 2.63\% improvement in the pageview rate through online experiments. In the ONCE \cite{once}, proposed by Liu et al., showed that using large-scale language models such as LLaMa \cite{llama} can improve news recommendation performance.

Several neural recommendation models use different architectures from the model shown in Figure \ref{fig:newsrec_overview}. For instance, Zhang et al. introduced UNBERT \cite{unbert}, which feeds a candidate news article and a list of previously viewed news articles into a single language model to predict click-through rates. Another notable work by Zhang et al. \cite{prompt4nr} is Prompt4NR, a news recommendation approach using prompt learning with PLMs.

In summary, various methods with neural network-based models have been proposed for news recommendation. The use of PLMs shows particularly high performance; however, PLMs lack sufficient knowledge to interpret category names. Thus, a space exists to enhance the performance of news recommendations by adding detailed category information, which inspires us to adopt LLM-generated category descriptions as additional features.

\subsection{Use of LLM-Generated Texts}
The use of text generated by LLMs, which have acquired extensive knowledge on a wide range of topics through pre-training, has improved the performance in various natural language processing tasks, such as text classification and question answering \cite{qagen,cupl,gpt3mix}.

Yoo et al. \cite{gpt3mix} proposed GPT3Mix, which applies GPT-3-based \cite{gpt3} data augmentation to text classification tasks by generating additional training examples. For question answering tasks, Liu et al. \cite{qagen} used knowledge expansion using GPT-3 to augment the context and improve the performance of the question answering model. Pratt et al. \cite{cupl} introduced CuPL, a method that leverages image captions generated by LLMs for zero-shot image classification using CLIP \cite{clip}. They automatically generated image captions using LLMs and used them as text inputs for CLIP.

These studies demonstrated the effectiveness of utilizing texts generated by LLMs in various tasks. Inspired by this, we aims to improve news recommendation performance by generating news category descriptions using LLMs.

\begin{figure*}[t]
    \includegraphics[width=\linewidth]{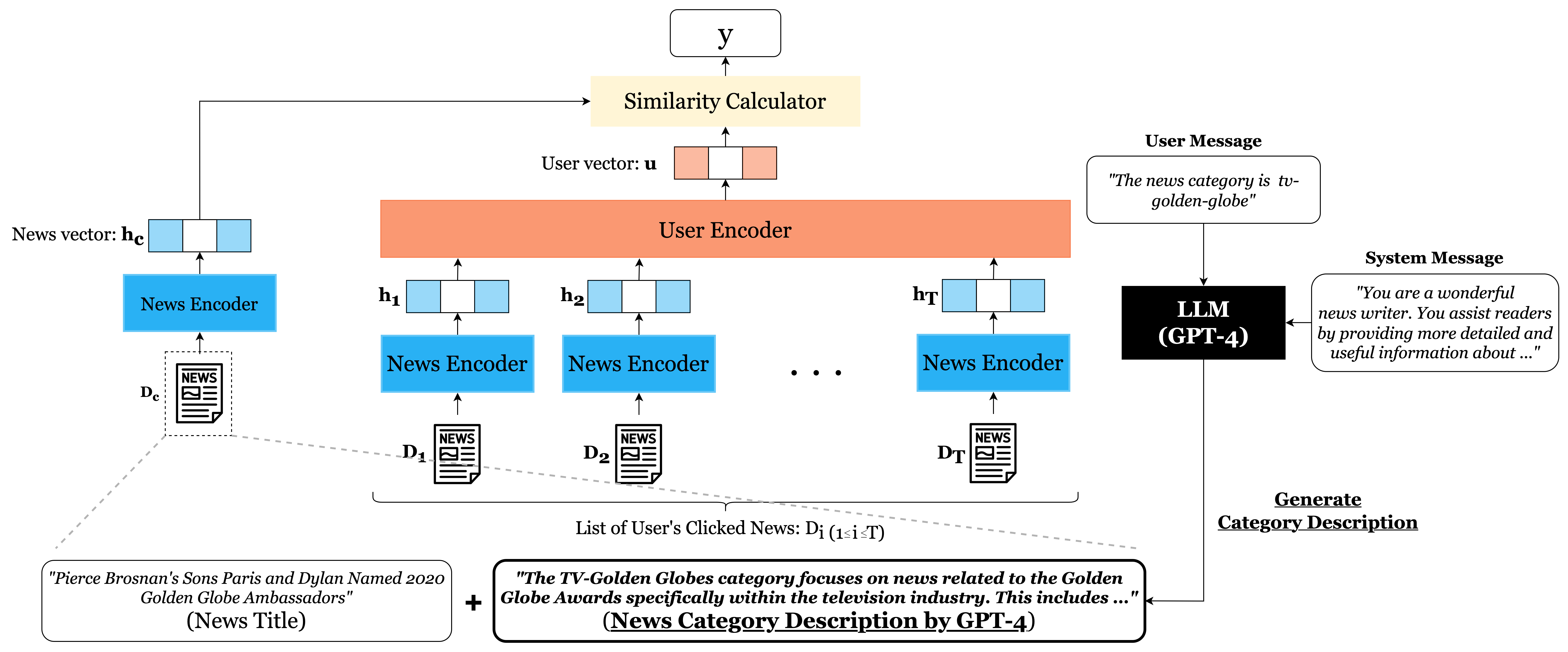}
    \caption{Overview of our proposed method}
    \label{fig:method_overview}
\end{figure*}

\section{Proposed Method}\label{sec:method}

Figure \ref{fig:method_overview} shows our proposed novel news recommendation method enhanced by LLM-generated category descriptions, consisting of two steps:  1) automatic generation of news category descriptions, and 2) integration of the generated category descriptions with the recommendation model as additional input features.

\subsection{Generation of Category Descriptions}\label{subsec:categoryGen}

In this step, we generate the descriptions for news categories using an LLM without any manual effort. We employ GPT-4 \cite{gpt4} as the LLM for generating category descriptions, as it has demonstrated high performance across various domains and tasks \cite{sparkGpt4}.

\begin{figure}[t]
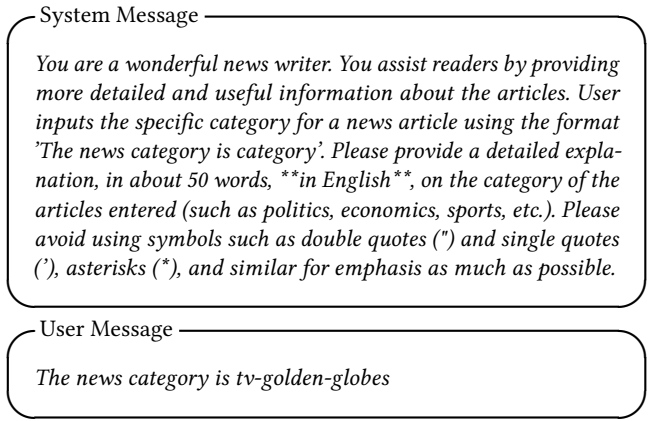

    \begin{itembox}[l]{System Message}
        \textit{You are a wonderful news writer. You assist readers by providing more detailed and useful information about the articles. User inputs the specific category for a news article using the format 'The news category is {category}'. Please provide a detailed explanation, in about 50 words, **in English**, on the category of the articles entered (such as politics, economics, sports, etc.). Please avoid using symbols such as double quotes (") and single quotes ('), asterisks (*), and similar for emphasis as much as possible.}
    \end{itembox}
    \begin{itembox}[l]{User Message}
        \textit{The news category is tv-golden-globes}
    \end{itembox}
    \caption{Prompt to generate a category description for \textit{tv-golden-glove} as an example.}
    \label{fig:prompt}
\end{figure}

Figure \ref{fig:prompt} shows the prompts for the LLM to generate descriptions for news categories. Although the prompts need to be prepared manually, they do not require any manual effort once they have been prepared.

\begin{figure}[t]
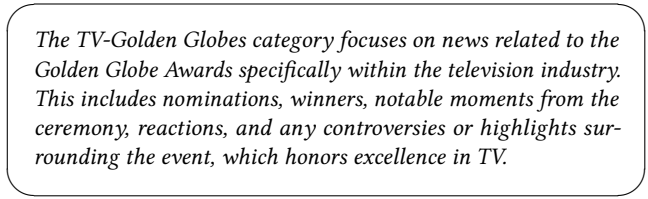

    \begin{screen}
        \textit{The TV-Golden Globes category focuses on news related to the Golden Globe Awards specifically within the television industry. This includes nominations, winners, notable moments from the ceremony, reactions, and any controversies or highlights surrounding the event, which honors excellence in TV.}
    \end{screen}
    \caption{Category description for \textit{tv-golden-globes} via GPT-4}
    \label{fig:outputexample}
\end{figure}

Figure \ref{fig:outputexample} shows a specific example of the generated news category descriptions for the \textit{tv-golden-globes} category, one of the categories that appeared in the MIND dataset \cite{mind}. The average word count of the generated category descriptions is 57.7 for all 270 news categories within the MIND dataset used in our experiment.

\subsection{Integration of Category Descriptions into News Recommendation Models}\label{subsec:catRecInput}

After generating the category descriptions, we train the news recommendation model by leveraging the generated descriptions. Let $D_{title}$ denote the news title text and $D_{desc}$ represent the generated category description. During the inference phase of the recommendation model, we concatenate $D_{title}$ and $D_{desc}$ using the special \textit{SEP} token from BERT \cite{bert}, and feed the resulting text into the news encoder.

\section{Experiments and Results}\label{sec:exp}

In this section, we describe our experiments. First, we present the dataset, evaluation metrics, baselines, and hyperparameters used in our experiments. Second, we present the results and then discuss the effectiveness of the proposed method.

\subsection{Experimental Setting}
We used the MIND \cite{mind} dataset, a widely used dataset for the news recommendation field. The MIND dataset, whose statistics are shown in Table \ref{tab:mind_statistic}, was constructed from the news click records of the Microsoft News\footnote{https://news.microsoft.com} service between October 12, 2019 and November 22, 2019.

\begin{table}[b]
    \centering
    \caption{MIND dataset \cite{mind}}
    \begin{tabular}{cccc} \hline
        \#User & \#News & \#Impression & \#Click \\
        94,057 & 65,238 & 230,117      & 347,727 \\ \hline
    \end{tabular}
    \label{tab:mind_statistic}
\end{table}

\begin{table*}[t]
    \centering
    \caption{Comparison of different methods on the MIND dataset}
    \begin{tabular}{lllcccc}
        \hline
        Recommendation Model              & PLM                         & Method                                       & AUC            & MRR            & nDCG@5         & nDCG@10        \\
        \hline
        \multirow{6}{*}{NAML \cite{naml}} & \multirow{3}{*}{DistilBERT} & \textit{title only}                          & 0.675          & 0.292          & 0.317          & 0.384          \\
                                          &                             & \textit{title + template-based}              & 0.690          & 0.295          & 0.327          & 0.393          \\
                                          &                             & \textit{title + generate-description} (ours) & \textbf{0.713} & \textbf{0.326} & \textbf{0.363} & \textbf{0.425} \\ \cline{2-7}
                                          & \multirow{3}{*}{BERT}       & \textit{title only}                          & 0.700          & 0.318          & 0.350          & 0.414          \\
                                          &                             & \textit{title + template-based}              & 0.696          & 0.308          & 0.340          & 0.405          \\
                                          &                             & \textit{title + generate-description} (ours) & \textbf{0.707} & \textbf{0.322} & \textbf{0.357} & \textbf{0.420} \\
        \hline
        \multirow{6}{*}{NRMS \cite{nrms}} & \multirow{3}{*}{DistilBERT} & \textit{title only}                          & 0.674          & 0.297          & 0.322          & 0.387          \\
                                          &                             & \textit{title + template-based}              & 0.675          & 0.311          & 0.341          & 0.400          \\
                                          &                             & \textit{title + generate-description} (ours) & \textbf{0.707} & \textbf{0.324} & \textbf{0.359} & \textbf{0.422} \\ \cline{2-7}
                                          & \multirow{3}{*}{BERT}       & \textit{title only}                          & 0.689          & 0.306          & 0.336          & 0.400          \\
                                          &                             & \textit{title + template-based}              & 0.667          & 0.301          & 0.329          & 0.389          \\
                                          &                             & \textit{title + generate-description} (ours) & \textbf{0.706} & \textbf{0.320} & \textbf{0.355} & \textbf{0.418} \\ \hline
        \multirow{6}{*}{NPA \cite{npa}}   & \multirow{3}{*}{DistilBERT} & \textit{title only}                          & 0.700          & 0.311          & 0.344          & 0.408          \\
                                          &                             & \textit{title + template-based}              & 0.698          & 0.309          & 0.342          & 0.407          \\
                                          &                             & \textit{title + generate-description} (ours) & \textbf{0.707} & \textbf{0.319} & \textbf{0.354} & \textbf{0.417} \\ \cline{2-7}
                                          & \multirow{3}{*}{BERT}       & \textit{title only}                          & 0.689          & 0.301          & 0.332          & 0.398          \\
                                          &                             & \textit{title + template-based}              & 0.694          & 0.314          & 0.345          & 0.410          \\
                                          &                             & \textit{title + generate-description} (ours) & \textbf{0.710} & \textbf{0.324} & \textbf{0.360} & \textbf{0.422} \\ \hline
    \end{tabular}
    \label{tab:evaluation_result}
\end{table*}

We compared our proposed method with the following two baselines:
\begin{enumerate}
    \item \textit{title only}: Input only the title text into the news encoder without any category information.
    \item \textit{title + template-based}: Input the concatenated text of the title and the template-based sentence that follows the template: \textit{The news category is \{category\}} (e.g., \textit{The news category is tv-golden-globes}), using the \textit{SEP} token.
\end{enumerate}

To evaluate the effectiveness of our proposed method, we adopted three state-of-the-art content-based recommendation models: NAML \cite{naml}, NRMS \cite{nrms}, and NPA \cite{npa}. We used the pre-trained models DistilBERT-base (\textit{distilbert-base-uncased}) \cite{distilbert} and BERT-base (\textit{bert-base-uncased}) \cite{bert} for the news encoder component. AdamW \cite{adamw} was used as the optimization function with a learning rate of 1e-4, a batch size of 128, and 3 epochs. Following the previous studies \cite{mccm,npa,plmnr,prompt4nr}, the maximum number of recently viewed articles, which was used to capture user preference information, was set to 50. We set the number of negative samples in the training phase to 4. All experiments are conducted on a Tesla V100 GPU. For implementation, we used PyTorch v2.1.0 \cite{pytorch} and Transformers v4.35.0 \cite{transformers} libraries. As performance metrics, we used the average AUC, MRR, nDCG@5, and nDCG@10 for all impressions.

\subsection{Results}

Table \ref{tab:evaluation_result} summarizes the comparison of our proposed recommendation method with the baselines. The proposed method, \textit{title+generated-description}, demonstrates the highest performance compared with the baselines across all recommendation models and achieved up to 5.6\% and 5.8\% improvements in AUC compared with the \textit{title only} and \textit{title+template-based} baselines, respectively.

Furthermore, \textit{title + template-based} shows insignificant improvement in performance compared with the \textit{title only} in most cases. This result suggests that simply fitting the category name into a predefined template is ineffective for representing the category information. The results also indicate that for relatively small-scale language models like BERT, using only the category names as input does not provide sufficient information for the recommendation models to properly interpret the category information.

\section{Limitations}\label{sec:limit}

To explore the limitations, we performed a manual inspection of the category descriptions. Although GPT-4 generally produces high-quality descriptions, it occasionally fails to generate accurate descriptions.

Figure \ref{fig:output_tunedin} shows an example in which the generated description for the \textit{tunedin} category focuses on entertainment, music, and television. However, the actual article titles in Table \ref{tab:tunedin_news_titles} reflect a wider range of topics, including technology and trends. A more accurate description would be that the \textit{tunedin} category casually provides news on various topics such as entertainment, technology, and business, keeping readers updated on the latest trends. This discrepancy highlights the model's inability to accurately capture the category's broad scope.

This example, along with other similar instances, suggests that the model may struggle to generate accurate category descriptions when the input lacks sufficient background knowledge or context.

\begin{table}[t]
    \centering
    \caption{News titles for the \textit{tunedin} category}
    \begin{tabular}{c} \hline
        News title                                                             \\ \hline
        \textit{Coca-Cola released two limited-edition holiday flavored sodas} \\
        \textit{Google Maps and Waze may share certain features}               \\
        \textit{Daylight saving time ends: Get ready to 'fall back'}           \\ \hline
    \end{tabular}
    \label{tab:tunedin_news_titles}
\end{table}

\begin{figure}[t]
    \begin{screen}
        \textit{The category "tunedin" typically refers to news related to entertainment, media, and television. It often includes updates on TV shows, interviews with celebrities, insights into the music industry, and information on streaming services. This category keeps readers engaged with the latest trends and happenings in the world of entertainment.}
    \end{screen}
    \caption{Category description for the \textit{tunedin} category}
    \label{fig:output_tunedin}
\end{figure}

\section{Conclusion}\label{sec:conclusion}

In this study, we proposed a method to automatically generate news category descriptions using LLMs and incorporate them into news recommendation models. The experiments on the MIND dataset demonstrated that our method outperformed the baselines across all metrics for multiple models. Our main contribution is a novel approach to enhance news recommendation models' understanding of category information using LLMs. Our future work will include enhancing the recommendation performance by improving the generated descriptions.


\printbibliography

@article{gpt4,
      title={{GPT-4 Technical Report}}, 
      author={OpenAI},
      year={2023},      
      journal = {arXiv preprint arXiv:2303.08774 (2023)},
}

@article{roberta,
  author = {Liu, Yinhan and Ott, Myle and Goyal, Naman and Du, Jingfei and Joshi, Mandar and Chen, Danqi and Levy, Omer and Lewis, Mike and Zettlemoyer, Luke and Stoyanov, Veselin},
  title = {{RoBERTa: A Robustly Optimized BERT Pretraining Approach}},
  year = {2019},
  journal = {arXiv preprint arXiv:1907.11692},
}

@article{clip,
  author={Alec, Radford and Jong, Wook-Kim and Chris, Hallacy and Aditya, Ramesh and Gabriel, Goh and Sandhini, Agarwal and Girish, Sastry and Amanda, Askell and Pamela, Mishkin and Jack, Clark and Gretchen, Krueger and Ilya, Sutskever},
  title = {{Learning Transferable Visual Models From Natural Language Supervision}},
  year = {2021},
  journal = {arXiv preprint arXiv:1907.11692},
}

@article{once,
      title={{ONCE: Boosting Content-based Recommendation with Both Open- and Closed-source Large Language Models}}, 
      author={Qijiong Liu and Nuo Chen and Tetsuya Sakai and Xiao-Ming Wu},
      year={2023},
      journal = {arXiv preprint arXiv:2305.06566},      
}

@article{llama,
      title={{LLaMA: Open and Efficient Foundation Language Models}}, 
      author={Hugo Touvron and Thibaut Lavril and Gautier Izacard and Xavier Martinet and Marie-Anne Lachaux and Timothée Lacroix and Baptiste Rozière and Naman Goyal and Eric Hambro and Faisal Azhar and Aurelien Rodriguez and Armand Joulin and Edouard Grave and Guillaume Lample},
      year={2023},
      journal = {arXiv preprint arXiv:2302.13971},
}

@inproceedings{reduceCross,
author = {Shivaram, Karthik and Liu, Ping and Shapiro, Matthew and Bilgic, Mustafa and Culotta, Aron},
title = {{Reducing Cross-Topic Political Homogenization in Content-Based News Recommendation}},
year = {2022},
address = {Seattle, WA, USA},
booktitle = {Proceedings of the 16th ACM Conference on Recommender Systems},
pages = {220-228},
}

@article{nrecsurvey2021,
author = {Wu, Chuhan and Wu, Fangzhao and Huang, Yongfeng and Xie, Xing},
title = {{Personalized News Recommendation: Methods and Challenges}},
year = {2023},
volume = {41},
journal = {ACM Trans. Inf. Syst.},
pages = {1-50},
}

@article{nrecsurvey2023,
author = {Meng, Xiangfu and Huo, Hongjin and Zhang, Xiaoyan and Wang, Wanchun and Zhu, Jinxia},
title = {{A Survey of Personalized News Recommendation}},
year = {2023},
journal = {Data Science and
Engineering},
pages = {1–21},
}

@article{sparkGpt4,
      title={{Sparks of Artificial General Intelligence: Early experiments with GPT-4}}, 
      author={Sébastien Bubeck and Varun Chandrasekaran and Ronen Eldan and Johannes Gehrke and Eric Horvitz and Ece Kamar and Peter Lee and Yin Tat Lee and Yuanzhi Li and Scott Lundberg and Harsha Nori and Hamid Palangi and Marco Tulio Ribeiro and Yi Zhang},
      year={2023},
      journal = {arXiv preprint arXiv:2303.12712},
}

@article{distilbert,
      title={{DistilBERT, a distilled version of BERT: smaller, faster, cheaper and lighter}}, 
      author={Victor Sanh and Lysandre Debut and Julien Chaumond and Thomas Wolf},
      year={2020},
      journal = {arXiv preprint arXiv:1910.01108},
}

@article{adamw,
  author={Loshchilov, Ilya and Hutter, Frank},
  title={{Decoupled weight decay regularization}},
  year={2017},
  journal = {arXiv preprint arXiv:1711.05101},
}

@inproceedings{gpt3,
author = {Brown, Tom B. and Mann, Benjamin and Ryder, Nick and Subbiah, Melanie and Kaplan, Jared and Dhariwal, Prafulla and Neelakantan, Arvind and Shyam, Pranav and Sastry, Girish and Askell, Amanda and Agarwal, Sandhini and Herbert-Voss, Ariel and Krueger, Gretchen and Henighan, Tom and Child, Rewon and Ramesh, Aditya and Ziegler, Daniel M. and Wu, Jeffrey and Winter, Clemens and Hesse, Christopher and Chen, Mark and Sigler, Eric and Litwin, Mateusz and Gray, Scott and Chess, Benjamin and Clark, Jack and Berner, Christopher and McCandlish, Sam and Radford, Alec and Sutskever, Ilya and Amodei, Dario},
title = {{Language models are few-shot learners}},
year = {2020},
isbn = {9781713829546},
publisher = {Curran Associates Inc.},
address = {Vancouver, BC, Canada},
booktitle = {Proceedings of the 34th International Conference on Neural Information Processing Systems},
pages = {1877-1901},
}

@inproceedings{gpt3mix,
    title = {{GPT3Mix: Leveraging Large-scale Language Models for Text Augmentation}},
    author = "Yoo, Kang Min  and
      Park, Dongju  and
      Kang, Jaewook  and
      Lee, Sang-Woo  and
      Park, Woomyoung",
    booktitle = "Findings of the Association for Computational Linguistics: EMNLP 2021",
    year = "2021",
    address = "Punta Cana, Dominican Republic",
    pages = {2225--2239},
}

@inproceedings{qagen,
    title = "Generated Knowledge Prompting for Commonsense Reasoning",
    author = "Liu, Jiacheng  and
      Liu, Alisa  and
      Lu, Ximing  and
      Welleck, Sean  and
      West, Peter  and
      Le Bras, Ronan  and
      Choi, Yejin  and
      Hajishirzi, Hannaneh",
    booktitle = "Proceedings of the 60th Annual Meeting of the Association for Computational Linguistics (Volume 1: Long Papers)",
    year = "2022",
    address = "Dublin, Ireland",
    pages = {3154-3169},
}

@inproceedings{cupl,
  author={Pratt, Sarah and Covert, Ian and Liu, Rosanne and Farhadi, Ali},
  booktitle={2023 IEEE/CVF International Conference on Computer Vision (ICCV)}, 
  title={{What does a platypus look like? Generating customized prompts for zero-shot image classification}}, 
  year={2023},
  address={Paris, France},
  pages={15645-15655}
}

@inproceedings{mmrec,
author = {Wu, Chuhan and Wu, Fangzhao and Qi, Tao and Zhang, Chao and Huang, Yongfeng and Xu, Tong},
title = {{MM-Rec: Visiolinguistic Model Empowered Multimodal News Recommendation}},
year = {2022},
address = {Madrid, Spain},
booktitle = {Proceedings of the 45th International ACM SIGIR Conference on Research and Development in Information Retrieval},
pages = {2560-2564},
}

@inproceedings{prompt4nr,
author = {Zhang, Zizhuo and Wang, Bang},
title = {{Prompt Learning for News Recommendation}},
year = {2023},
booktitle = {{Proceedings of the 46th International ACM SIGIR Conference on Research and Development in Information Retrieval}},
address = {Taipei, Taiwan},
pages={227-237}
}

@inproceedings{unbert,
  title     = {{UNBERT: User-News Matching BERT for News Recommendation}},
  author    = {Zhang, Qi and Li, Jingjie and Jia, Qinglin and Wang, Chuyuan and Zhu, Jieming and Wang, Zhaowei and He, Xiuqiang},
  booktitle = {Proceedings of the Thirtieth International Joint Conference on Artificial Intelligence, {IJCAI-21}},
  address = {Virtual},
   year      = {2021},
  pages     = {3356-3362},
}

@inproceedings{nrms,
    title = {{Neural News Recommendation with Multi-Head Self-Attention}},
    author = "Wu, Chuhan  and
      Wu, Fangzhao  and
      Ge, Suyu  and
      Qi, Tao  and
      Huang, Yongfeng  and
      Xie, Xing",
    booktitle = "Proceedings of the 2019 Conference on Empirical Methods in Natural Language Processing and the 9th International Joint Conference on Natural Language Processing (EMNLP-IJCNLP)",
    year = "2019",
    address = "Hong Kong, China",
    pages = {6389-6394},
}

@inproceedings{npa,
author = {Wu, Chuhan and Wu, Fangzhao and An, Mingxiao and Huang, Jianqiang and Huang, Yongfeng and Xie, Xing},
title = {{NPA: Neural News Recommendation with Personalized Attention}},
year = {2019},
booktitle = {{Proceedings of the 25th ACM SIGKDD International Conference on Knowledge Discovery \& Data Mining}},
pages = {2576-2584},
address = {Anchorage, AK, USA},
}

@inproceedings{lstur,
    title = {{Neural News Recommendation with Long- and Short-term User Representations}},
    author = "An, Mingxiao  and
      Wu, Fangzhao  and
      Wu, Chuhan  and
      Zhang, Kun  and
      Liu, Zheng  and
      Xie, Xing",
    booktitle = "Proceedings of the 57th Annual Meeting of the Association for Computational Linguistics",
    year = "2019",
    address = "Florence, Italy",
    pages = {336-345},
}

@inproceedings{plmnr,
author = {Wu, Chuhan and Wu, Fangzhao and Qi, Tao and Huang, Yongfeng},
title = {{Empowering News Recommendation with Pre-Trained Language Models}},
address = {Virtual},
booktitle = {Proceedings of the 44th International ACM SIGIR Conference on Research and Development in Information Retrieval},
year = {2021},
pages = {1652-1656},
}

@inproceedings{naml,
author = {Wu, Chuhan and Wu, Fangzhao and An, Mingxiao and Huang, Jianqiang and Huang, Yongfeng and Xie, Xing},
title = {{Neural news recommendation with attentive multi-view learning}},
year = {2019},
pages = {3863–3869},
address = {Macao, China},
booktitle = {Proceedings of the Twenty-Eighth International Joint Conference on Artificial Intelligence, {IJCAI-19}}
}

@inproceedings{mind,
    title = {{MIND: A Large-scale Dataset for News Recommendation}},
    author = "Wu, Fangzhao  and
      Qiao, Ying  and
      Chen, Jiun-Hung  and
      Wu, Chuhan  and
      Qi, Tao  and
      Lian, Jianxun  and
      Liu, Danyang  and
      Xie, Xing  and
      Gao, Jianfeng  and
      Wu, Winnie  and
      Zhou, Ming",
    booktitle = "Proceedings of the 58th Annual Meeting of the Association for Computational Linguistics",
    year = "2020",
    address = "Virtual",
    pages = {3597-3606},
}

@inproceedings{bert,
    title = {{BERT: Pre-training of Deep Bidirectional Transformers for Language Understanding}},
    author = "Devlin, Jacob  and
      Chang, Ming-Wei  and
      Lee, Kenton  and
      Toutanova, Kristina",
    booktitle = "Proceedings of the 2019 Conference of the North {A}merican Chapter of the Association for Computational Linguistics: Human Language Technologies",
    year = "2019",
    address = "Minneapolis, Minnesota",
    pages = {4171-4186},
}

@inproceedings{mccm,
author = {Wang, Jingkun and Jiang, Yongtao and Li, Haochen and Zhao, Wen},
title = {{Improving News Recommendation with Channel-Wise Dynamic Representations and Contrastive User Modeling}},
year = "2023",
address = "Singapore",
booktitle = "Proceedings of the Sixteenth ACM International Conference on Web Search and Data Mining",
pages = {562-570},
}

@inproceedings{transformers,
    title = {{Transformers: State-of-the-Art Natural Language Processing}},
    author = "Wolf, Thomas  and
      Debut, Lysandre  and
      Sanh, Victor  and
      Chaumond, Julien  and
      Delangue, Clement  and
      Moi, Anthony  and
      Cistac, Pierric  and
      Rault, Tim  and
      Louf, Remi  and
      Funtowicz, Morgan  and
      Davison, Joe  and
      Shleifer, Sam  and
      von Platen, Patrick  and
      Ma, Clara  and
      Jernite, Yacine  and
      Plu, Julien  and
      Xu, Canwen  and
      Le Scao, Teven  and
      Gugger, Sylvain  and
      Drame, Mariama  and
      Lhoest, Quentin  and
      Rush, Alexander",
    booktitle = "Proceedings of the 2020 Conference on Empirical Methods in Natural Language Processing: System Demonstrations",
    year = "2020",
    address = "Virtual",
    pages = {38-45},
}

@inproceedings{pytorch,
author = {Paszke, Adam and Gross, Sam and Massa, Francisco and Lerer, Adam and Bradbury, James and Chanan, Gregory and Killeen, Trevor and Lin, Zeming and Gimelshein, Natalia and Antiga, Luca and Desmaison, Alban and K\"{o}pf, Andreas and Yang, Edward and DeVito, Zach and Raison, Martin and Tejani, Alykhan and Chilamkurthy, Sasank and Steiner, Benoit and Fang, Lu and Bai, Junjie and Chintala, Soumith},
title = {{PyTorch: An Imperative Style, High-Performance Deep Learning Library}},
year = {2019},
address = {Red Hook, NY, USA},
booktitle = {Proceedings of the 33rd International Conference on Neural Information Processing Systems},
pages={8024-8035}
}


\end{document}